\documentclass{article}
\usepackage{float}
\usepackage{amsmath}
\usepackage{PRIMEarxiv}
\usepackage{makecell}
\usepackage[utf8]{inputenc} 
\usepackage[T1]{fontenc}    
\usepackage{hyperref}       
\usepackage{url}            
\usepackage{booktabs}       
\usepackage{amsfonts}       
\usepackage{nicefrac}       
\usepackage{microtype}      
\usepackage{lipsum}
\usepackage{fancyhdr}       
\usepackage{graphicx}       
\graphicspath{{media/}} 
\usepackage{xcolor}
\usepackage{authblk}
\usepackage{hyperref}
\usepackage{multirow} 
\usepackage{longtable} 
\usepackage{enumitem}
\usepackage{multirow}
\usepackage{array}
\usepackage{verbatim}
\begin{document}

\title{Evaluating the Ability of Large Language Models to Identify Adherence to CONSORT Reporting Guidelines in Randomized Controlled Trials: A Methodological Evaluation Study}
\author[1,†]{Zhichao He}
\author[2,†]{Mouxiao Bian}
\author[1,†]{Jianhong Zhu}
\author[2]{Jiayuan Chen}
\author[1,3]{Yunqiu Wang}
\author[1]{Wenxia Zhao}
\author[2]{Tianbin Li}
\author[2]{Bing Han}
\author[2,*]{Jie Xu}
\author[1,*]{Junyan Wu}
\affil[1]{\textit{
    SUN YAT-SEN MEMORIAL HOSPITAL \\
    Guangdong, China
}}
\affil[2]{\textit{
    Shanghai Artificial Intelligence Laboratory\\
    Shanghai, China
}}
\affil[3]{\textit{
    Imperial College London\\
    London, United Kingdom
}}
\affil[4]{\textit{
   University of Washington\\
    Washington, USA
}}
 
\footnotetext[1]{†These authors contributed equally.}
\footnotetext[2]{*Correspondence: 
Junyan Wu(wujunyan@mail.sysu.edu.cn),
Jie Xu (xujie@pjlab.org.cn)
}

\maketitle
\begin{abstract}
\textbf{\textbf{Background:}} The Consolidated Standards of Reporting Trials (CONSORT) statement is the global benchmark for transparent and high-quality reporting of randomized controlled trials (RCTs). Manual verification of CONSORT adherence is a laborious, time-intensive process that constitutes a significant bottleneck in peer review and evidence synthesis. While Large Language Models (LLMs) have demonstrated transformative potential in natural language understanding, their proficiency in the nuanced, high-stakes task of methodological assessment remains unverified.

\textbf{\textbf{Objective:}} This study aimed to systematically evaluate the accuracy and reliability of contemporary LLMs in identifying the adherence of published RCTs to the CONSORT 2010 statement under a zero-shot setting.

\textbf{\textbf{Methods:}} We constructed a gold-standard dataset of 150 published RCTs spanning diverse medical specialties. Each article was independently assessed against 37 CONSORT sub-items by two trained methodologists, with discrepancies resolved by a senior arbitrator to establish a consensus label ("Compliant," "Non-Compliant," or "Not Applicable"). A suite of 16 leading LLMs, including GPT, Gemini, Claude, and Qwen series, were evaluated. The primary outcome was the macro-averaged F1-score for the three-class classification task, supplemented by item-wise performance metrics and qualitative error analysis.

\textbf{\textbf{Results:}} Overall model performance was modest. The top-performing models, Gemini-2.5-Flash and DeepSeek-R1, achieved nearly identical macro F1-scores of 0.634 and Cohen's Kappa coefficients of 0.280 and 0.282, respectively, indicating only "fair" agreement with expert consensus. A striking performance disparity was observed across classes: while most models could identify "Compliant" items with high accuracy (F1-score > 0.850), they struggled profoundly with identifying "Non-Compliant" and "Not Applicable" items, where F1-scores rarely exceeded 0.400. Notably, some high-profile models like GPT-4o underperformed, achieving a macro F1-score of only 0.521. A primary systematic error was the misclassification of "Not Reported" as "Not Applicable," particularly for items concerning changes to trial methods or outcomes.

\textbf{\textbf{Conclusion:}} LLMs show potential as preliminary screening assistants for CONSORT checks, capably identifying well-reported items. However, their current inability to reliably detect reporting omissions or methodological flaws ("Non-Compliant" items) makes them unsuitable for replacing human expertise in the critical appraisal of trial quality. Their utility is presently confined to augmenting, not automating, the work of researchers, editors, and systematic reviewers.

\end{abstract}


\section{Introduction}

Randomized controlled trials (RCTs) represent the pinnacle of study design for evaluating the efficacy and safety of health interventions, forming the bedrock of evidence-based medicine \cite{guyatt1992evidence}\cite{moher2010consort}. The validity of their conclusions and their utility in clinical decision-making and health policy are, however, critically dependent on the transparent, complete, and accurate reporting of their design, conduct, and analysis \cite{ioannidis2005most}\cite{chalmers2009avoidable}. Deficiencies in reporting can obscure methodological flaws, introduce bias, and impede the replication of research, ultimately leading to a waste of valuable research resources\cite{glasziou2014reducing}\cite{moher2016increasing}.

To address these pervasive issues, the Consolidated Standards of Reporting Trials (CONSORT) statement was developed. This evidence-based set of recommendations, provides a 25-item checklist designed to guide authors in reporting RCTs, thereby improving their quality and transparency \cite{schulz2010consort}. The endorsement of CONSORT by hundreds of leading medical journals has been associated with modest to significant improvements in reporting quality \cite{turner2012consolidated}\cite{turner2012does}, yet adherence remains suboptimal across the medical literature\cite{chhapolareporting}\cite{han2009impact}.

Manually verifying adherence to CONSORT guidelines is a standard but demanding component of the peer-review process for journals and a foundational step in conducting systematic reviews and meta-analyses\cite{page2021prisma}\cite{cumpston2019updated}. This process is not only time-consuming and resource-intensive but also requires substantial methodological expertise and can be subject to inter-rater variability\cite{2007The}\cite{2020Impact}. The increasing volume of published research exacerbates this challenge, creating a significant bottleneck in the evidence synthesis pipeline\cite{2010Seventy}.

The recent advent of large language models (LLMs) such as OpenAI's GPT series, Google's Gemini, and Anthropic's Claude has revolutionized the field of artificial intelligence\cite{brown2020language}\cite{team2023gemini}\cite{martinez2025generative}. Their sophisticated capabilities in understanding, summarizing, and reasoning over complex text have opened new frontiers for their application in medicine and scientific research \cite{thirunavukarasu2023large}. From assisting in clinical documentation and drafting manuscripts to accelerating drug discovery, LLMs are poised to reshape the research landscape \cite{singhal2023large}\cite{madani2023large}\cite{galli2025large}. The potential to automate the assessment of scientific literature for methodological rigor is particularly tantalizing, promising to enhance efficiency and consistency \cite{yan2024practical}\cite{lin2025roles}.

However, assessing CONSORT adherence transcends simple keyword searching or text summarization. It demands a nuanced understanding of complex epidemiological and biostatistical concepts such as allocation concealment, blinding, and intention-to-treat analysis\cite{altman2001revised}\cite{schulz1995empirical}. It requires the ability to infer information, identify omissions, and critically evaluate the adequacy of reported details\cite{higgins2011cochrane}. The extent to which current LLMs possess this specialized reasoning capability, particularly without domain-specific fine-tuning (i.e., in a "zero-shot" context), is unknown.

This study, therefore, seeks to address a critical question: How accurately and reliably can state-of-the-art LLMs identify adherence to CONSORT reporting guidelines in published RCTs compared to a gold standard of human expert consensus? We hypothesized that LLMs would perform well on straightforward, descriptive items but would struggle with items requiring deep methodological inference. By establishing a rigorous benchmark—RCTBench, this research aims to provide comprehensive evidence on the capabilities and limitations of LLMs for this critical task, informing their potential role as an assistive tool for researchers, peer reviewers, and systematic review teams.

\section{Methods}
\subsection{Study Design}
This was a methodological evaluation study designed to benchmark the performance of multiple LLMs against a human-expert-derived gold standard for the task of CONSORT 2010 guideline adherence checking.

\subsection{Gold Standard Dataset Construction}

A dataset of 150 full-text articles of RCTs published between 2020 and 2025 was compiled. The articles were sourced from PubMed and the Cochrane Central Register of Controlled Trials (CENTRAL), ensuring a diverse sample across various medical and surgical specialties, journal impact factors, trial designs (e.g., parallel, crossover), and sample sizes.

A rigorous annotation process was implemented to create the gold standard. Ten methodologists with postgraduate training in epidemiology or biostatistics served as expert annotators. Each article was independently assessed by two annotators against the 37 sub-items of the CONSORT 2010 checklist. For each item, a three-way classification was made: "Compliant" (the item was adequately reported), "Non-Compliant" (the item was inadequately reported or omitted), or "Not Applicable" (the item was not relevant to the specific trial design, e.g., Item 11 on blinding in an open-label trial). All discrepancies between the two primary annotators were reviewed by a third, senior methodologist who arbitrated a final consensus label. For each "Non-Compliant" or "Not Applicable" judgment, annotators provided a concise, standardized justification and answer point.

\subsection{Large Language Models and Evaluation Task}

We evaluated a panel of 16 prominent LLMs representing the state-of-the-art as of early 2024. This included models from OpenAI (GPT-4o, GPT-5 series), Google (Gemini series), Anthropic (Claude series), Alibaba (Qwen series), Meta (Llama-4 series), DeepSeek AI, and Mistral AI.
The evaluation was conducted in a zero-shot setting to assess the models' intrinsic capabilities without task-specific examples. A detailed prompt was engineered to instruct the models to act as an evidence-based medicine expert. The prompt required the models to read the full text of an RCT and output a structured JSON object containing two arrays: one for "non\_compliant\_items" and one for "not\_applicable\_items." Each entry in these arrays was required to include the CONSORT item number and a brief reason for the classification. Compliant items were to be omitted from the output for brevity and efficiency. This structured output format facilitated automated parsing and scoring, prompt can be found in appendix 2.

\subsection{Outcome Measures and Statistical Analysis}

The primary outcome measure was the macro-averaged F1-score (Macro-F1), which computes the F1-score for each class ("Compliant," "Non-Compliant," "Not Applicable") independently and then averages them, giving equal weight to each class..
\begin{equation}
\label{F1_score}
F_1 = 2 \times \frac{\text{Precision} \times \text{Recall}}{\text{Precision} + \text{Recall}}
\end{equation}

Where:\\
- Precision represents the proportion of samples predicted as positive that are actually positive(Equation \ref{precision}):
 \begin{equation} 
 \label{precision}
  \text{Precision} = \frac{TP}{TP + FP}
  \end{equation}
- Recall represents the proportion of actual positive samples that are correctly predicted as positive(Equation \ref{recall}):
\begin{equation}
 \label{recall}
  \text{Recall} = \frac{TP}{TP + FN}
  \end{equation}

In the above formulas, $TP$ stands for True Positives, $FP$ stands for False Positives, and $FN$ stands for False Negatives.
This metric is robust to class imbalance. Macro-averaged precision and recall were also calculated. Cohen's Kappa coefficient was used to measure the agreement between model predictions and the gold standard, accounting for chance agreement.
Secondary outcomes included item-wise F1-scores and class-specific performance metrics (precision, recall, F1) to identify specific areas of model strength and weakness. A qualitative error analysis was performed by categorizing the justifications provided by models for their incorrect predictions, particularly focusing on systematic error patterns. All statistical analyses were descriptive, with results presented in tables and figures.

\section{Results}
\subsection{Gold Standard Dataset Characteristics}

\begin{figure}
    \centering
    \includegraphics[width=1\linewidth]{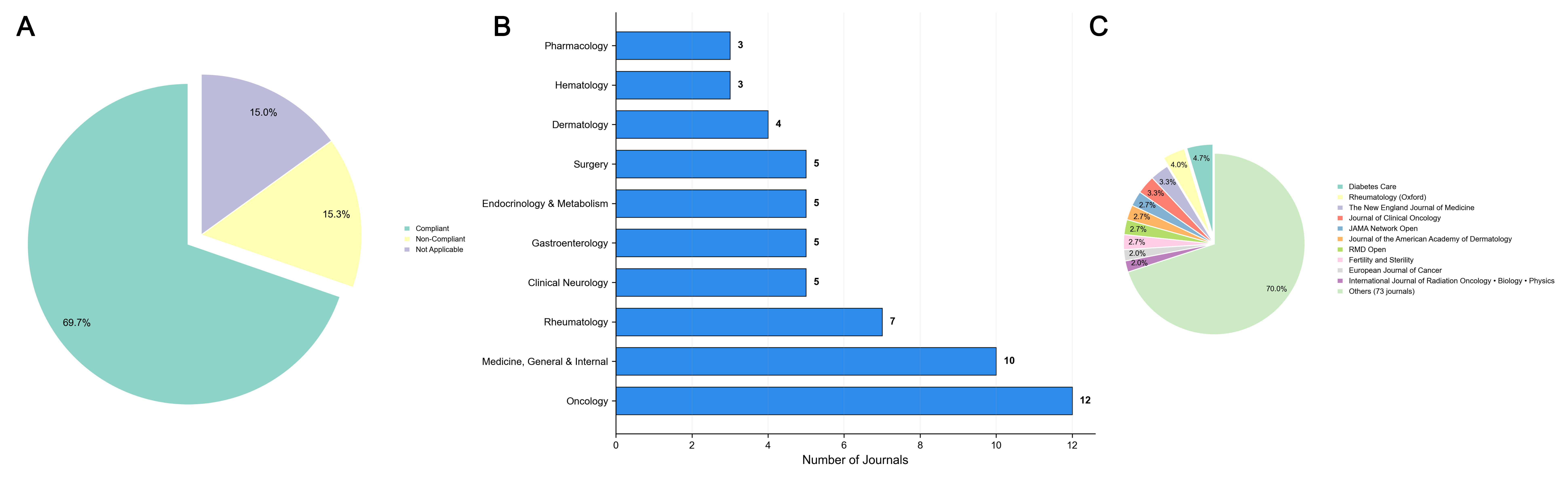}
    \caption{Characteristics of RCTBench}
    \label{fig:characteristic}
\end{figure}

RCTBench  consists of 150 RCTs comprising 5,550 assessable CONSORT items and contains articles from 83 journals, 38 disciplines (Figure \ref{fig:characteristic}) . Expert consensus classified 3,868 (69.7\%) as "Compliant," 847 (15.0\%) as "Non-Compliant," and 835 (15.3\%) as "Not Applicable." This distribution highlights the prevalence of reporting deficiencies and the necessity of handling "Not Applicable" cases in real-world scenarios. Items with the highest non-compliance rates were Item 10 (randomization implementation; 87.3\%), Item 9 (allocation concealment; 62.0\%), and Item 14a (recruitment dates; 56.0\%), indicating these are common areas of poor reporting.
\subsection{Overall Model Performance}

The performance of the 16 LLMs was modest and highly variable, with a clear distinction between identifying compliance and detecting non-compliance. As shown in Table \ref{tab:overall_erformance}, Gemini-2.5-Flash and DeepSeek-R1 were the top performers, achieving nearly identical macro F1-scores of 0.634. Their Cohen's Kappa values of 0.280 and 0.282, respectively, indicate only "fair" agreement with the gold standard.

A striking finding was the underperformance of several highly anticipated models. GPT-4o, for example, achieved a macro F1-score of only 0.521 and a Kappa of 0.097, indicating slight agreement. Models like Llama-4-Maverick and Llama-4-Scout performed at or below the level of chance, with negative Kappa coefficients.

\begin{table}
\centering
\caption{Overall Performance of Selected Large Language Models on CONSORT Adherence Classification}
\label{tab:overall_erformance}
\begin{tabular}{ l  l  l  l  l }
\toprule
\textbf{\textbf{Model Name}} & \textbf{\textbf{Macro Precision}} & \textbf{\textbf{Macro Recall}} & \textbf{\textbf{Macro F1-Score}} & \textbf{\textbf{Cohen's Kappa}} \\
\midrule
Gemini-2.5-Flash & 0.699 & 0.496 & 0.634 & 0.280 \\

DeepSeek-R1 & 0.700 & 0.496 & 0.634 & 0.282 \\

GPT-5mini & 0.714 & 0.514 & 0.633 & 0.277 \\

Gemini-2.5-Pro & 0.670 & 0.470 & 0.630 & 0.279 \\

GPT-5 & 0.677 & 0.553 & 0.617 & 0.262 \\

gpt-oss-20b & 0.691 & 0.485 & 0.613 & 0.251 \\

Claude Sonnet4-20250514 & 0.645 & 0.423 & 0.581 & 0.172 \\

DeepSeek-V3.1 & 0.588 & 0.441 & 0.577 & 0.166 \\

DeepSeek-V3 & 0.590 & 0.423 & 0.564 & 0.135 \\

Qwen3\_32b & 0.585 & 0.399 & 0.557 & 0.122 \\

Mistral-Small-3.1-24B-Instruct-2503 & 0.559 & 0.423 & 0.553 & 0.121 \\

Qwen3-235B-A22B & 0.531 & 0.408 & 0.542 & 0.097 \\

GPT-4o & 0.549 & 0.400 & 0.521 & 0.097 \\

Llama-4-Scout-17B-16E & 0.459 & 0.354 & 0.482 & -0.016 \\

Qwen2.5-72B-instruct & 0.558 & 0.376 & 0.476 & 0.019 \\

Llama-4-Maverick-17B-128E-Instruct & 0.520 & 0.377 & 0.473 & -0.009 \\
\bottomrule

\end{tabular}

\end{table}

\subsection{Class-Specific Performance and Item-wise Analysis}

\begin{figure}
    \centering
    \includegraphics[width=0.5\linewidth]{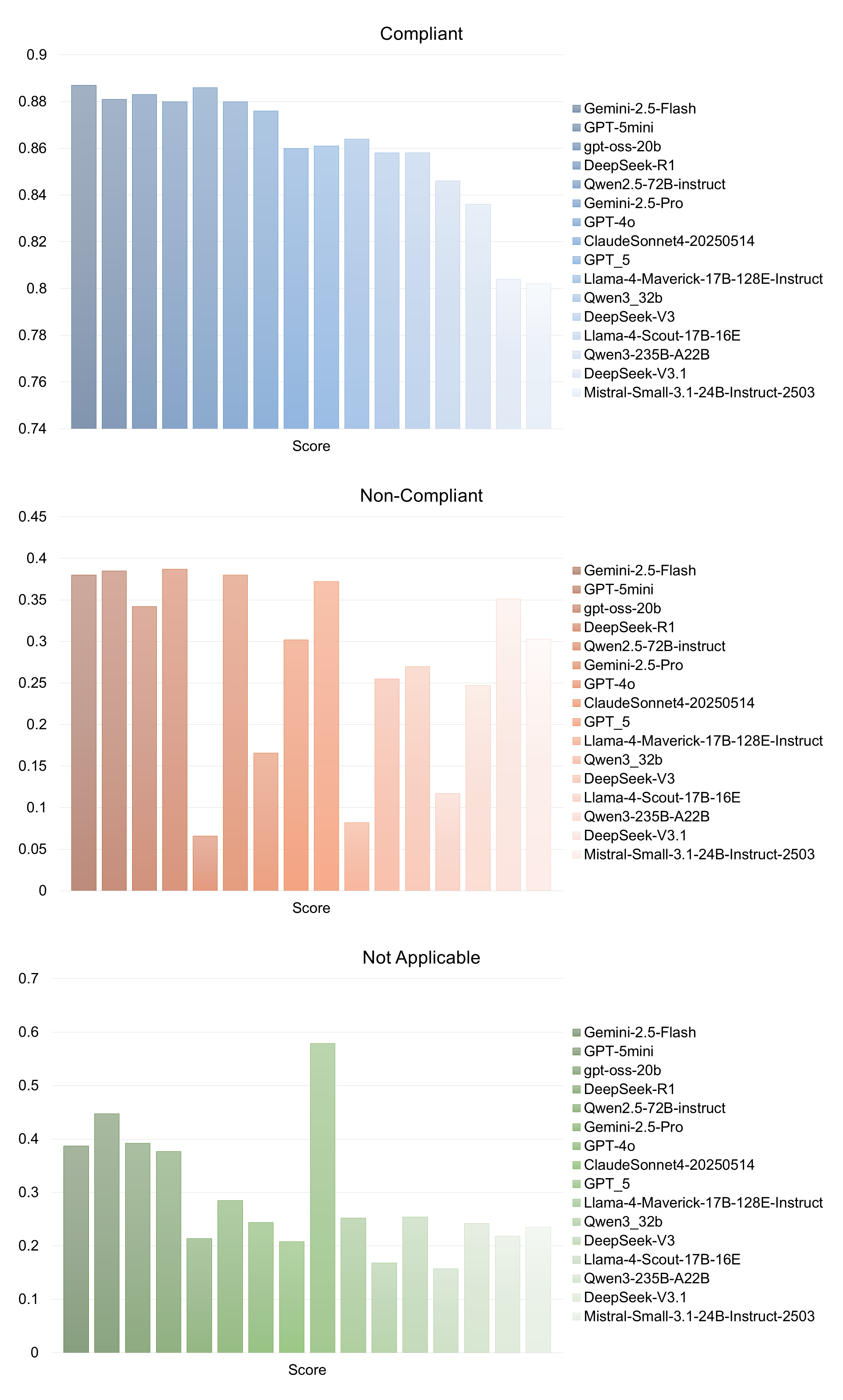}
    \caption{Performance Dichotomy (F1-Score) by Class for Top and Notable Models}
    \label{fig:F1Score}
\end{figure}

The modest macro-F1 scores obscure a critical performance dichotomy between classes (Figure \ref{fig:F1Score}). All models were proficient at identifying "Compliant" items. Top models like Gemini-2.5-Flash and GPT-5mini achieved F1-scores of 0.887 and 0.881 for this class, respectively, driven by high recall (>0.900 for many). This suggests models can effectively recognize when information is present.

However, performance collapsed when attempting to identify "Non-Compliant" or "Not Applicable" items. For the crucial "Non-Compliant" class, the highest F1-score was only 0.385 (GPT-5mini), with most models scoring well below 0.350. The F1-score for the "Not Applicable" class was similarly poor, with the best model (GPT-5) reaching 0.579 and many falling below 0.400. GPT-4o, for instance, scored a mere 0.166 F1 for "Non-Compliant" and 0.244 for "Not Applicable," indicating it missed the vast majority of these cases.

This pattern explains why models with very high overall accuracy (e.g., Gemini-2.5-Flash at 80.6\%) still have low Kappa and Macro F1 scores: they default to predicting the majority class ("Compliant"), thereby failing at the primary task of flagging deficiencies.

\subsection{Qualitative Error Analysis}

Qualitative analysis of errors provided a clear explanation for the poor quantitative results. The most frequent systematic error was the conflation of "information not reported" with "not applicable." For items requiring a statement about whether a procedure occurred (e.g., Item 3b: changes to methods; Item 7b: interim analyses; Item 14b: trial discontinuation), models consistently provided justifications like, "The article does not mention any important changes to the methods," to support an incorrect "Not Applicable" classification. The correct judgment, according to CONSORT, is "Non-Compliant" because the omission itself is a reporting failure.

Another major error involved a shallow interpretation of methodology. Models frequently cited text describing random sequence generation (Item 8a) as evidence for adequate allocation concealment (Item 9), failing to distinguish these distinct concepts. This indicates a reliance on keyword proximity rather than a procedural understanding of bias mitigation techniques.

\section{Discussion}

This comprehensive evaluation reveals that while LLMs possess some capability for automating CONSORT checks, their performance is modest and their limitations are profound. Our central finding is a stark performance dichotomy: models are adept at confirming the presence of reported information but are largely incapable of reliably identifying reporting omissions or methodological flaws. The overall performance, with the best models achieving only "fair" agreement ($Kappa \approx  $0.28), falls far short of the threshold required for autonomous use in high-stakes scientific appraisal.

The results challenge the optimistic assumption that larger or more recent models are universally better for specialized tasks. The surprising underperformance of a prominent model like GPT-4o (Macro F1 = 0.521) compared to Gemini-2.5-Flash (Macro F1 = 0.634) highlights that model architecture, training data composition, and fine-tuning for instruction-following can lead to idiosyncratic strengths and weaknesses. GPT-4o's very low recall for non-compliant and not-applicable items suggests it may be "over-calibrated" to avoid making definitive negative claims in the absence of explicit negative evidence, a behavior that is detrimental in a prescriptive checking task.

The near-total failure to identify "Non-Compliant" items (F1 < 0.400) is the most critical limitation uncovered. The primary value of a CONSORT checking tool is not to confirm what is present, but to flag what is missing or inadequately reported, as these omissions are central to assessing risk of bias\cite{altman2001revised} \cite{wood2008empirical}. The models' tendency to default to the "Compliant" class makes them unreliable for this core function. This aligns with findings from other domains showing that LLMs struggle with tasks requiring "negative evidence" or reasoning about omissions\cite{bommasani2021opportunities}\cite{dhuliawala2023chain}.

Our qualitative analysis pinpoints the root cause: models lack a deep, normative understanding of reporting guidelines. They operate on a descriptive logic ("what is in the text?") rather than a prescriptive one ("what \textit{\textit{should be}} in the text according to the standard?"). The recurring error of misclassifying "not reported" as "not applicable" is a direct consequence of this logical gap \cite{moher2010consort}. This suggests that simply scaling up existing model architectures may be insufficient; new training paradigms that instill normative reasoning are needed\cite{dhuliawala2023chain}\cite{shneiderman2020human}.

Despite these limitations, our findings do not dismiss the potential of LLMs. Their high recall for compliant items suggests they can be used as effective "rule-out" or preliminary screening tools. A workflow where an LLM first performs a rapid scan to identify and clear obviously well-reported items could allow human experts to focus their limited time on the more ambiguous and methodologically complex aspects of a paper \cite{2020Impact}\cite{spillias2024human}. This human-in-the-loop model appears to be the most viable application for the current generation of LLMs \cite{amershi2019guidelines}.

This study has several strengths, including its rigorous gold-standard, comprehensive model testing, and detailed error analysis. Limitations include the zero-shot setting, which may understate the potential of fine-tuned or few-shot models \cite{gururangan2020don}\cite{ouyang2022training}; the exclusion of information from figures and tables, which future multimodal models may address \cite{hopewell2011reporting}\cite{li2023blip}; and the rapidly evolving nature of LLM technology \cite{shamsabadi2024large}.

\section{Conclusion}

Large language models, in their current state, are not reliable autonomous agents for assessing CONSORT reporting guideline adherence. While they can efficiently identify adequately reported information, their profound inability to detect reporting omissions or methodological flaws renders them unsafe for standalone use in critical appraisal. Their immediate value lies in a more modest role: as assistive tools in a human-supervised workflow, helping to streamline the initial, more tedious aspects of the review process. The significant gap between human expert performance and current AI capabilities underscores the enduring necessity of deep methodological expertise in safeguarding the integrity of clinical evidence.

\section{\textbf{Data Availability Statement}}
The dataset supporting the findings of this study has been made publicly available through the MedBench repository. It can be accessed at \href{https://medbench.opencompass.org.cn/home}{https://medbench.opencompass.org.cn/home}.
\bibliography{references.bib} 

@article{guyatt1992evidence,
  title={Evidence-based medicine: a new approach to teaching the practice of medicine},
  author={Guyatt, Gordon and Cairns, John and Churchill, David and Cook, Deborah and Haynes, Brian and Hirsh, Jack and Irvine, Jan and Levine, Mark and Levine, Mitchell and Nishikawa, Jim and others},
  journal={jama},
  volume={268},
  number={17},
  pages={2420--2425},
  year={1992},
  publisher={American Medical Association}
}

@article{moher2010consort,
  title={CONSORT 2010 explanation and elaboration: updated guidelines for reporting parallel group randomised trials},
  author={Moher, David and Hopewell, Sally and Schulz, Kenneth F and Montori, Victor and G{\o}tzsche, Peter C and Devereaux, Philip J and Elbourne, Diana and Egger, Matthias and Altman, Douglas G},
  journal={Bmj},
  volume={340},
  year={2010},
  publisher={British Medical Journal Publishing Group}
}

@article{ioannidis2005most,
  title={Why most published research findings are false},
  author={Ioannidis, John PA},
  journal={PLoS medicine},
  volume={2},
  number={8},
  pages={e124},
  year={2005},
  publisher={Public Library of Science}
}

@article{chalmers2009avoidable,
  title={Avoidable waste in the production and reporting of research evidence},
  author={Chalmers, Iain and Glasziou, Paul},
  journal={The Lancet},
  volume={374},
  number={9683},
  pages={86--89},
  year={2009},
  publisher={Elsevier}
}

@article{glasziou2014reducing,
  title={Reducing waste from incomplete or unusable reports of biomedical research},
  author={Glasziou, Paul and Altman, Douglas G and Bossuyt, Patrick and Boutron, Isabelle and Clarke, Mike and Julious, Steven and Michie, Susan and Moher, David and Wager, Elizabeth},
  journal={The Lancet},
  volume={383},
  number={9913},
  pages={267--276},
  year={2014},
  publisher={Elsevier}
}

@article{moher2016increasing,
  title={Increasing value and reducing waste in biomedical research: who's listening?},
  author={Moher, David and Glasziou, Paul and Chalmers, Iain and Nasser, Mona and Bossuyt, Patrick MM and Korevaar, Dani{\"e}l A and Graham, Ian D and Ravaud, Philippe and Boutron, Isabelle},
  journal={The Lancet},
  volume={387},
  number={10027},
  pages={1573--1586},
  year={2016},
  publisher={Elsevier}
}

@article{schulz2010consort,
  title={CONSORT 2010 statement: updated guidelines for reporting parallel group randomised trials},
  author={Schulz, Kenneth F and Altman, Douglas G and Moher, David and Consort Group and others},
  journal={Journal of clinical epidemiology},
  volume={63},
  number={8},
  pages={834--840},
  year={2010},
  publisher={Elsevier}
}

@article{turner2012consolidated,
  title={Consolidated standards of reporting trials (CONSORT) and the completeness of reporting of randomised controlled trials (RCTs) published in medical journals},
  author={Turner, Lucy and Shamseer, Larissa and Altman, Douglas G and Weeks, Laura and Peters, Jodi and Kober, Thilo and Dias, Sofia and Schulz, Kenneth F and Plint, Amy C and Moher, David},
  journal={Cochrane database of systematic reviews},
  number={11},
  year={2012},
  publisher={John Wiley \& Sons, Ltd}
}

@article{turner2012does,
  title={Does use of the CONSORT Statement impact the completeness of reporting of randomised controlled trials published in medical journals? A Cochrane reviewa},
  author={Turner, Lucy and Shamseer, Larissa and Altman, Douglas G and Schulz, Kenneth F and Moher, David},
  journal={Systematic reviews},
  volume={1},
  number={1},
  pages={60},
  year={2012},
  publisher={Springer}
}

@misc{chhapolareporting,
  title={Reporting quality of trial abstracts-improved yet suboptimal: A systematic review and meta-analysis. J Evid Based Med. 2018; 11 (2): 89-94},
  author={Chhapola, V and Tiwari, S and Brar, R and Kanwal, SK}
}

@article{han2009impact,
  title={The impact of the CONSORT statement on reporting of randomized clinical trials in psychiatry},
  author={Han, Changsu and Kwak, Kyung-phil and Marks, David M and Pae, Chi-Un and Wu, Li-Tzy and Bhatia, Kamal S and Masand, Prakash S and Patkar, Ashwin A},
  journal={Contemporary Clinical Trials},
  volume={30},
  number={2},
  pages={116--122},
  year={2009},
  publisher={Elsevier}
}

@article{page2021prisma,
  title={The PRISMA 2020 statement: an updated guideline for reporting systematic reviews},
  author={Page, Matthew J and McKenzie, Joanne E and Bossuyt, Patrick M and Boutron, Isabelle and Hoffmann, Tammy C and Mulrow, Cynthia D and Shamseer, Larissa and Tetzlaff, Jennifer M and Akl, Elie A and Brennan, Sue E and others},
  journal={bmj},
  volume={372},
  year={2021},
  publisher={British Medical Journal Publishing Group}
}

@article{cumpston2019updated,
  title={Updated guidance for trusted systematic reviews: a new edition of the Cochrane Handbook for Systematic Reviews of Interventions},
  author={Cumpston, Miranda and Li, Tianjing and Page, Matthew J and Chandler, Jacqueline and Welch, Vivian A and Higgins, Julian PT and Thomas, James},
  journal={The Cochrane database of systematic reviews},
  volume={2019},
  number={10},
  pages={ED000142},
  year={2019}
}

@article{2007The,
  title={The quality of reporting of orthopaedic randomized trials with use of a checklist for nonpharmacological therapies.},
  author={ Chan, Simon  and  Bhandari, Mohit },
  journal={journal of bone \& joint surgery american volume},
  volume={89},
  number={9},
  pages={1970-8},
  year={2007},
}

@article{2020Impact,
  title={Impact of blinding on estimated treatment effects in randomised clinical trials: meta-epidemiological study},
  author={ Clayton, Gemma L  and  Jones, Hayley E  and  Boutron, Isabelle  and  Laursen, David L T  and  Olsen, Mette F  and  Ravaud, Philippe  and  Savovi, Jelena  and  Sterne, Jonathan A C  and Asbjrn Hróbjartsson and  Jrgensen, Lars },
  journal={RMD Open},
  volume={368},
  pages={-},
  year={2020},
}

@article{2010Seventy,
  title={Seventy-Five Trials and Eleven Systematic Reviews a Day: How Will We Ever Keep Up?},
  author={ Bastian, Hilda  and  Glasziou, Paul  and  Chalmers, Iain },
  journal={Plos Medicine},
  volume={7},
  number={9},
  pages={e1000326},
  year={2010},
}

@article{brown2020language,
  title={Language models are few-shot learners},
  author={Brown, Tom and Mann, Benjamin and Ryder, Nick and Subbiah, Melanie and Kaplan, Jared D and Dhariwal, Prafulla and Neelakantan, Arvind and Shyam, Pranav and Sastry, Girish and Askell, Amanda and others},
  journal={Advances in neural information processing systems},
  volume={33},
  pages={1877--1901},
  year={2020}
}

@article{team2023gemini,
  title={Gemini: a family of highly capable multimodal models},
  author={Team, Gemini and Anil, Rohan and Borgeaud, Sebastian and Alayrac, Jean-Baptiste and Yu, Jiahui and Soricut, Radu and Schalkwyk, Johan and Dai, Andrew M and Hauth, Anja and Millican, Katie and others},
  journal={arXiv preprint arXiv:2312.11805
        },
  year={2023}
}

@article{martinez2025generative,
  title={Generative artificial intelligence-supported pentesting: a comparison between claude opus, gpt-4, and copilot},
  author={Mart{\'\i}nez, Antonio L{\'o}pez and Cano, Alejandro and Ruiz-Mart{\'\i}nez, Antonio},
  journal={arXiv preprint arXiv:2501.06963 
        },
  year={2025}
}

@article{thirunavukarasu2023large,
  title={Large language models in medicine},
  author={Thirunavukarasu, Arun James and Ting, Darren Shu Jeng and Elangovan, Kabilan and Gutierrez, Laura and Tan, Ting Fang and Ting, Daniel Shu Wei},
  journal={Nature medicine},
  volume={29},
  number={8},
  pages={1930--1940},
  year={2023},
  publisher={Nature Publishing Group US New York}
}

@article{singhal2023large,
  title={Large language models encode clinical knowledge},
  author={Singhal, Karan and Azizi, Shekoofeh and Tu, Tao and Mahdavi, S Sara and Wei, Jason and Chung, Hyung Won and Scales, Nathan and Tanwani, Ajay and Cole-Lewis, Heather and Pfohl, Stephen and others},
  journal={Nature},
  volume={620},
  number={7972},
  pages={172--180},
  year={2023},
  publisher={Nature Publishing Group}
}

@article{galli2025large,
  title={Large Language Models in Systematic Review Screening: Opportunities, Challenges, and Methodological Considerations},
  author={Galli, Carlo and Gavrilova, Anna V and Calciolari, Elena},
  journal={Information},
  volume={16},
  number={5},
  pages={378},
  year={2025},
  publisher={MDPI}
}

@article{madani2023large,
  title={Large language models generate functional protein sequences across diverse families},
  author={Madani, Ali and Krause, Ben and Greene, Eric R and Subramanian, Subu and Mohr, Benjamin P and Holton, James M and Olmos Jr, Jose Luis and Xiong, Caiming and Sun, Zachary Z and Socher, Richard and others},
  journal={Nature biotechnology},
  volume={41},
  number={8},
  pages={1099--1106},
  year={2023},
  publisher={Nature Publishing Group US New York}
}

@article{yan2024practical,
  title={Practical and ethical challenges of large language models in education: A systematic scoping review},
  author={Yan, Lixiang and Sha, Lele and Zhao, Linxuan and Li, Yuheng and Martinez-Maldonado, Roberto and Chen, Guanliang and Li, Xinyu and Jin, Yueqiao and Ga{\v{s}}evi{\'c}, Dragan},
  journal={British Journal of Educational Technology},
  volume={55},
  number={1},
  pages={90--112},
  year={2024},
  publisher={Wiley Online Library}
}

@article{lin2025roles,
  title={Roles and potential of large language models in healthcare: a comprehensive review},
  author={Lin, Chihung and Kuo, Chang-Fu},
  journal={Biomedical Journal},
  pages={100868},
  year={2025},
  publisher={Elsevier}
}

@article{altman2001revised,
  title={The revised CONSORT statement for reporting randomized trials: explanation and elaboration},
  author={Altman, Douglas G and Schulz, Kenneth F and Moher, David and Egger, Matthias and Davidoff, Frank and Elbourne, Diana and G{\o}tzsche, Peter C and Lang, Thomas and Consort Group},
  journal={Annals of internal medicine},
  volume={134},
  number={8},
  pages={663--694},
  year={2001},
  publisher={American College of Physicians}
}

@article{schulz1995empirical,
  title={Empirical evidence of bias: dimensions of methodological quality associated with estimates of treatment effects in controlled trials},
  author={Schulz, Kenneth F and Chalmers, Iain and Hayes, Richard J and Altman, Douglas G},
  journal={Jama},
  volume={273},
  number={5},
  pages={408--412},
  year={1995},
  publisher={American Medical Association}
}

@article{higgins2011cochrane,
  title={The Cochrane Collaboration’s tool for assessing risk of bias in randomised trials},
  author={Higgins, Julian PT and Altman, Douglas G and G{\o}tzsche, Peter C and J{\"u}ni, Peter and Moher, David and Oxman, Andrew D and Savovi{\'c}, Jelena and Schulz, Kenneth F and Weeks, Laura and Sterne, Jonathan AC},
  journal={bmj},
  volume={343},
  year={2011},
  publisher={British Medical Journal Publishing Group}
}

@article{wood2008empirical,
  title={Empirical evidence of bias in treatment effect estimates in controlled trials with different interventions and outcomes: meta-epidemiological study},
  author={Wood, Lesley and Egger, Matthias and Gluud, Lise Lotte and Schulz, Kenneth F and J{\"u}ni, Peter and Altman, Douglas G and Gluud, Christian and Martin, Richard M and Wood, Anthony JG and Sterne, Jonathan AC},
  journal={bmj},
  volume={336},
  number={7644},
  pages={601--605},
  year={2008},
  publisher={British Medical Journal Publishing Group}
}

@article{bommasani2021opportunities,
  title={On the opportunities and risks of foundation models},
  author={Bommasani, Rishi},
  journal={arXiv preprint arXiv:2108.07258 
        },
  year={2021}
}

@article{dhuliawala2023chain,
  title={Chain-of-verification reduces hallucination in large language models},
  author={Dhuliawala, Shehzaad and Komeili, Mojtaba and Xu, Jing and Raileanu, Roberta and Li, Xian and Celikyilmaz, Asli and Weston, Jason},
  journal={arXiv preprint arXiv:2309.11495},
  year={2023}
}

@article{shneiderman2020human,
  title={Human-centered artificial intelligence: Reliable, safe \& trustworthy},
  author={Shneiderman, Ben},
  journal={International Journal of Human--Computer Interaction},
  volume={36},
  number={6},
  pages={495--504},
  year={2020},
  publisher={Taylor \& Francis}
}

@article{spillias2024human,
  title={Human-AI collaboration to identify literature for evidence synthesis},
  author={Spillias, Scott and Tuohy, Paris and Andreotta, Matthew and Annand-Jones, Ruby and Boschetti, Fabio and Cvitanovic, Christopher and Duggan, Joseph and Fulton, Elisabeth A and Karcher, Denis B and Paris, Cecile and others},
  journal={Cell Reports Sustainability},
  volume={1},
  number={7},
  year={2024},
  publisher={Elsevier}
}

@inproceedings{amershi2019guidelines,
  title={Guidelines for human-AI interaction},
  author={Amershi, Saleema and Weld, Dan and Vorvoreanu, Mihaela and Fourney, Adam and Nushi, Besmira and Collisson, Penny and Suh, Jina and Iqbal, Shamsi and Bennett, Paul N and Inkpen, Kori and others},
  booktitle={Proceedings of the 2019 chi conference on human factors in computing systems},
  pages={1--13},
  year={2019}
}

@article{gururangan2020don,
  title={Don't stop pretraining: Adapt language models to domains and tasks},
  author={Gururangan, Suchin and Marasovi{\'c}, Ana and Swayamdipta, Swabha and Lo, Kyle and Beltagy, Iz and Downey, Doug and Smith, Noah A},
  journal={arXiv preprint arXiv:2004.10964
        
        },
  year={2020}
}

@article{ouyang2022training,
  title={Training language models to follow instructions with human feedback},
  author={Ouyang, Long and Wu, Jeffrey and Jiang, Xu and Almeida, Diogo and Wainwright, Carroll and Mishkin, Pamela and Zhang, Chong and Agarwal, Sandhini and Slama, Katarina and Ray, Alex and others},
  journal={Advances in neural information processing systems},
  volume={35},
  pages={27730--27744},
  year={2022}
}

@article{hopewell2011reporting,
  title={Reporting of participant flow diagrams in published reports of randomized trials},
  author={Hopewell, Sally and Hirst, Allison and Collins, Gary S and Mallett, Sue and Yu, Ly-Mee and Altman, Douglas G},
  journal={Trials},
  volume={12},
  number={1},
  pages={253},
  year={2011},
  publisher={Springer}
}

@inproceedings{li2023blip,
  title={Blip-2: Bootstrapping language-image pre-training with frozen image encoders and large language models},
  author={Li, Junnan and Li, Dongxu and Savarese, Silvio and Hoi, Steven},
  booktitle={International conference on machine learning},
  pages={19730--19742},
  year={2023},
  organization={PMLR}
}

@article{shamsabadi2024large,
  title={Large language models for scientific information extraction: An empirical study for virology},
  author={Shamsabadi, Mahsa and D'Souza, Jennifer and Auer, S{\"o}ren},
  journal={arXiv preprint arXiv:2401.10040
        },
  year={2024}
}
\bibliographystyle{IEEEtran} 
\section{Appendix}
\subsection{CONSORT 2010 Checklist}
\renewcommand{\arraystretch}{1.3}
\label{tab:consort}
\begin{longtable}{p{4.5cm} l p{7cm}} 
\hline
\textbf{Section/Topic} & \textbf{Item No} & \textbf{Checklist item} \\
\hline
\endhead 
\hline
\endfoot
\hline
\endlastfoot

\multirow{2}{=}{Title and abstract} & 1a & Identification as a randomised trial in the title \\
 & 1b & Structured summary of trial design, methods, results, and conclusions (for specific guidance see CONSORT for abstracts) \\
\hline
\multirow{2}{=}{Introduction: Background and objectives} & 2a & Scientific background and explanation of rationale \\
 & 2b & Specific objectives or hypotheses \\
\hline
\multirow{2}{=}{Methods: Trial design} & 3a & Description of trial design (such as parallel, factorial) including allocation ratio \\
 & 3b & Important changes to methods after trial commencement (such as eligibility criteria), with reasons \\
\hline
\multirow{2}{=}{Methods: Participants} & 4a & Eligibility criteria for participants \\
 & 4b & Settings and locations where the data were collected \\
\hline
\multirow{1}{=}{Methods: Interventions} & 5 & The interventions for each group with sufficient details to allow replication, including how and when they were actually administered \\
\hline
\multirow{2}{=}{Methods: Outcomes} & 6a & Completely defined pre-specified primary and secondary outcome measures, including how and when they were assessed \\
 & 6b & Any changes to trial outcomes after the trial commenced, with reasons \\
\hline
\multirow{2}{=}{Methods: Sample size} & 7a & How sample size was determined \\
 & 7b & When applicable, explanation of any interim analyses and stopping guidelines \\
\hline
\multirow{2}{=}{Methods: Randomisation: Sequence generation} & 8a & Method used to generate the random allocation sequence \\
 & 8b & Type of randomisation; details of any restriction (such as blocking and block size) \\
\hline
\multirow{1}{=}{Methods: Allocation concealment mechanism} & 9 & Mechanism used to implement the random allocation sequence (such as sequentially numbered containers), describing any steps taken to conceal the sequence until interventions were assigned \\
\hline
\multirow{1}{=}{Methods: Implementation} & 10 & Who generated the random allocation sequence, who enrolled participants, and who assigned participants to interventions \\
\hline
\multirow{2}{=}{Methods: Blinding} & 11a & If done, who was blinded after assignment to interventions (for example, participants, care providers, those assessing outcomes) and how \\
 & 11b & If relevant, description of the similarity of interventions \\
\hline
\multirow{2}{=}{Methods: Statistical methods} & 12a & Statistical methods used to compare groups for primary and secondary outcomes \\
 & 12b & Methods for additional analyses, such as subgroup analyses and adjusted analyses \\
\hline
\multirow{2}{=}{Results: Participant flow (a diagram is strongly recommended)} & 13a & For each group, the numbers of participants who were randomly assigned, received intended treatment, and were analysed for the primary outcome \\
 & 13b & For each group, losses and exclusions after randomisation, together with reasons \\
\hline
\multirow{2}{=}{Results: Recruitment} & 14a & Dates defining the periods of recruitment and follow-up \\
 & 14b & Why the trial ended or was stopped \\
\hline
\multirow{1}{=}{Results: Baseline data} & 15 & A table showing baseline demographic and clinical characteristics for each group \\
\hline
\multirow{1}{=}{Results: Numbers analysed} & 16 & For each group, number of participants (denominator) included in each analysis and whether the analysis was by original assigned groups \\
\hline
\multirow{2}{=}{Results: Outcomes and estimation} & 17a & For each primary and secondary outcome, results for each group, and the estimated effect size and its precision (such as 95\% confidence interval) \\
 & 17b & For binary outcomes, presentation of both absolute and relative effect sizes is recommended \\
\hline
\multirow{1}{=}{Results: Ancillary analyses} & 18 & Results of any other analyses performed, including subgroup analyses and adjusted analyses, distinguishing pre-specified from exploratory \\
\hline
\multirow{1}{=}{Results: Harms} & 19 & All important harms or unintended effects in each group (for specific guidance see CONSORT for harms) \\
\hline
\multirow{1}{=}{Discussion: Limitations} & 20 & Trial limitations, addressing sources of potential bias, imprecision, and, if relevant, multiplicity of analyses \\
\hline
\multirow{1}{=}{Discussion: Generalisability} & 21 & Generalisability (external validity, applicability) of the trial findings \\
\hline
\multirow{1}{=}{Discussion: Interpretation} & 22 & Interpretation consistent with results, balancing benefits and harms, and considering other relevant evidence \\
\hline
\multirow{1}{=}{Other information: Registration} & 23 & Registration number and name of trial registry \\
\hline
\multirow{1}{=}{Other information: Protocol} & 24 & Where the full trial protocol can be accessed, if available \\
\hline
\multirow{1}{=}{Other information: Funding} & 25 & Sources of funding and other support (such as supply of drugs), role of funders \\

\end{longtable}

\subsection{Prompt  designed for RCTBench}
Act as a rigorous systematic review and evidence-based medicine expert, and evaluate the provided clinical randomized controlled trial (RCT) literature strictly in accordance with the CONSORT Statement.
\paragraph{Your tasks:}
\begin{enumerate}[label=\arabic*., leftmargin=*]
    \item Carefully read the full text of the literature provided by the user.
    \item Verify each of the following CONSORT checklist items one by one.
    \item For each item, make one of the three judgments below:
    \begin{itemize}[label=--, leftmargin=*]
        \item \textbf{Compliant}: The literature clearly reports all requirements of the item.
        \item \textbf{Non-Compliant}: The literature does not report or reports incompletely.
        \item \textbf{Not Applicable}: The item is not applicable to this study (e.g., items 11a and 11b are "Not Applicable" in an open-label trial without blinding).
    \end{itemize}
    \item You need to output a structured JSON object containing two keys: \texttt{non\_compliant\_items} and \texttt{not\_applicable\_items}.
    \item \textbf{For items judged as "Non-Compliant"}, fill their numbers and reasons into the \texttt{non\_compliant\_items} array.
    \item \textbf{For items judged as "Not Applicable"}, fill their numbers and brief reasons into the \texttt{not\_applicable\_items} array.
    \item \textbf{Do not output} items judged as "Compliant".
\end{enumerate}

\paragraph{Output format requirements:}
The output must be pure, directly parsable JSON without any additional Markdown formatting or explanatory text..The JSON object structure must strictly follow the example below:

\begin{verbatim}
{
  "non_compliant_items": [
    {"item_number": "3b", "reason": "Failed to state any modifications to eligibility 
    criteria after the trial started and the reasons for such modifications"},
    {"item_number": "9", "reason": "Failed to describe the mechanism of allocation 
    concealment (e.g., type of container used) and implementation steps"}
  ],
  "not_applicable_items": [
    {"item_number": "11b", "reason": "This trial adopted an open-label design 
    without blinding, so there is no need to describe the similarity of interventions."}
  ]
}
\end{verbatim}
- If an array is empty (i.e., no "Non-Compliant" or "Not Applicable" items), the array shall be an empty array \texttt{[]}.
\paragraph{CONSORT Checklist:}
...(CONSORT 2010 Checklist,table \ref{tab:consort})...

\end{document}